\documentclass[sigconf]{acmart}

\usepackage{booktabs} % For formal tables
\usepackage{fixme}
\usepackage{color}

\setcopyright{rightsretained}
\copyrightyear{2017} 
\acmYear{2017} 
\setcopyright{acmlicensed}
\acmConference{FDG'17}{August 14-17, 2017}{Hyannis, MA, USA}\acmPrice{15.00}\acmDOI{10.1145/3102071.3110576}
\acmISBN{978-1-4503-5319-9/17/08}

\begin{document}

\title{Automatic Mapping of NES Games with Mappy}

\author{Joseph Osborn}
\orcid{0000-0003-0025-9525}
\affiliation{\institution{Computational Media}
 \institution{University of California at Santa Cruz}
 \city{Santa Cruz} 
 \state{California} 
 \postcode{95064}
}
\email{jcosborn@soe.ucsc.edu}

\author{Adam Summerville}
\affiliation{\institution{Computational Media}
  \institution{University of California at Santa Cruz }
  \streetaddress{1156 High St.}
  \city{ Santa Cruz} 
  \state{CA} 
  \postcode{95064 }
}
\email{asummerv@ucsc.edu }

\author{Michael Mateas}
\affiliation{
  \institution{Computational Media}
  \institution{University of California at Santa Cruz}
  \streetaddress{1156 High Street}
  \city{Santa Cruz} 
  \state{CA} 
  \postcode{95064}
}
\email{michaelm@soe.ucsc.edu}

% The default list of authors is too long for headers}
%\renewcommand{\shortauthors}{B. Trovato et al.}

\begin{abstract}
  Game maps are useful for human players, general-game-playing agents, and data-driven procedural content generation. 
  These maps are generally made by hand-assembling manually-created screenshots of game levels.
  Besides being tedious and error-prone, this approach requires additional effort for each new game and level to be mapped.
  The results can still be hard for humans or computational systems to make use of, privileging visual appearance over semantic information.
  We describe a software system, \emph{Mappy}, that produces a good approximation of a linked map of rooms given a Nintendo Entertainment System game program and a sequence of button inputs exploring its world.
  In addition to visual maps, \emph{Mappy} outputs grids of tiles (and how they change over time), positions of non-tile objects, clusters of similar rooms that might in fact be the same room, and a set of links between these rooms.
  We believe this is a necessary step towards developing larger corpora of high-quality semantically-annotated maps for PCG via machine learning and other applications. 
\end{abstract}

%
% The code below should be generated by the tool at
% http://dl.acm.org/ccs.cfm
% Please copy and paste the code instead of the example below. 
%


% We no longer use \terms command
%\terms{Theory}


\maketitle

\section{Introduction}

\begin{figure}[hbtp!]
\centering
\includegraphics[width=0.48\textwidth]{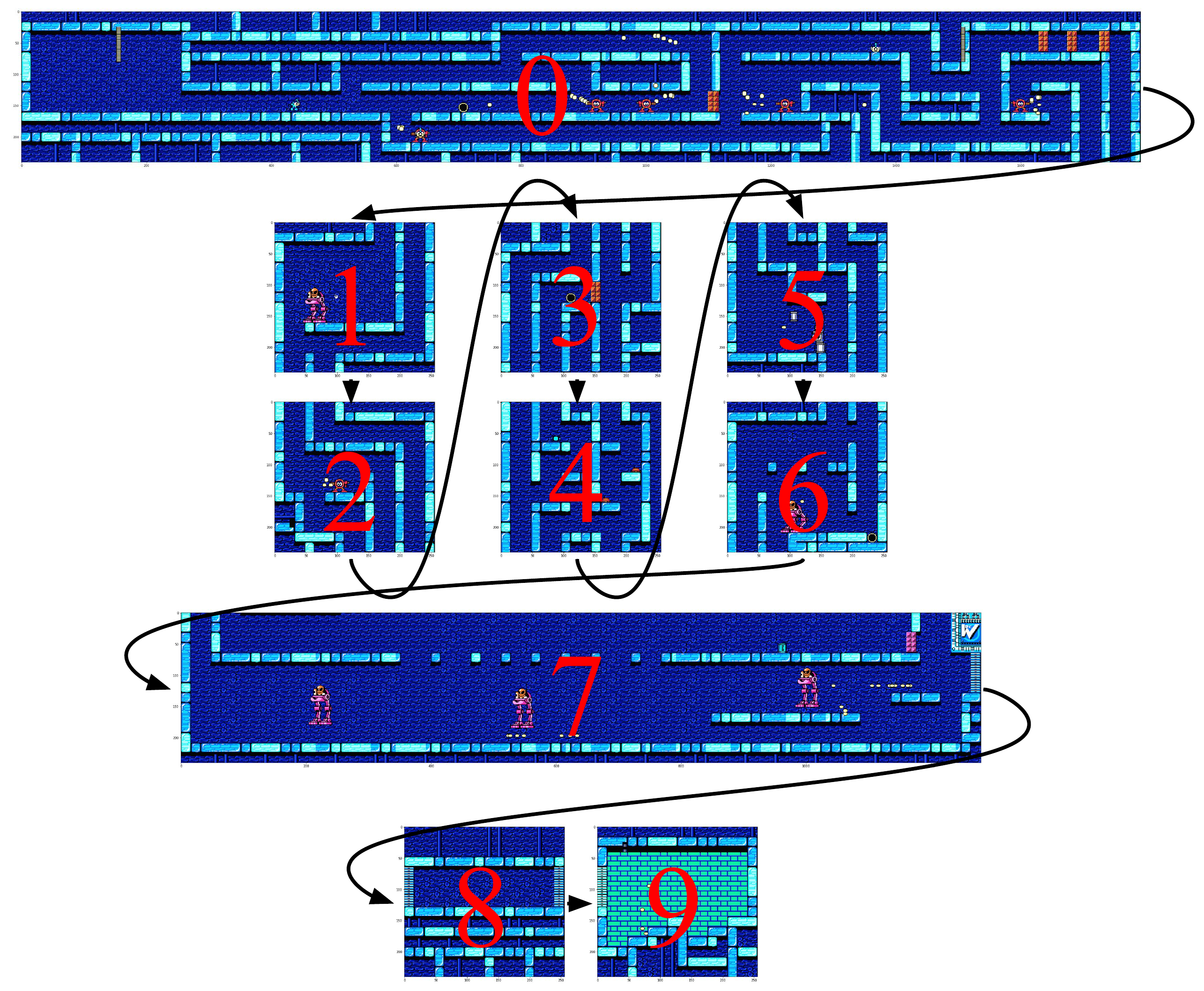}

\caption{Bright Man's stage from \textit{Mega Man 2}. \textit{Mappy} handles rooms and arrangements of arbitrary size.}
\label{fig:MM2}
\end{figure}

The production of game maps---both of local, scrolling rooms and the global structure that links those rooms together---is of interest to the general game-playing audience and has applications to general videogame AI and to data-driven procedural content generation.
Human players need, create, and make use of maps to improve their play.
Players track locations in their heads, draw informal maps, and in some cases capture screenshots and meticulously compose them into game world atlases.

Many games, including action-adventure games like \emph{Metroid} or \emph{The Legend of Zelda}, essentially compose together two distinct games: an action game of dodging and attacking played at sixty frames per second, and an adventure puzzle game in which the game map forms a graph search problem.
In this latter game, players engage in activities such as retrieving items from one room to open another room's exit, returning to suspicious locations to see what new opportunities have opened up, and collecting supplies before proceeding to a boss monster.
AI game-playing agents that do not remember both high-level map structures and sufficient low-level details of each room cannot hope to formulate plans like these on their own.
Clean, complete maps of game worlds (including the ways in which they might change due to player actions) are also integral to research in automatically learning game rules~\cite{summerville2017what} and to data-driven procedural content generation~\cite{DBLP:journals/corr/SummervilleSGHH17}.

In this paper we describe \emph{Mappy}, a partial solution for automatically mapping certain classes of Nintendo Entertainment System (NES) games (see Fig.~\ref{fig:MM2} for an example of a map made by \emph{Mappy}).
We choose to work with the NES because it is extremely well-known, it has a large and diverse set of supported games, there are well-annotated corpora of recorded input sequences including speed-runs, efficient emulators are readily available, and the NES hardware exposes several useful high-level features which facilitate automatic mapping.
The techniques we detail here could readily be extended to other tile-based rendering systems including the Super NES. To apply our approach to other game platforms, software renderers would need to expose information about the tilemap being drawn for the lower-level feature extraction and the data representation to be applicable.
We also consider only games without modal menus or other non-spatial uses of tiles.
Specifically, we assume that a (given) sub-rectangle of the screen is \emph{scrolling} and all other regions of the screen can be safely ignored.

\section{Related Work}

\subsection{Manual Mapping}

While there are several online communities at work producing full maps of games, this process is laborious and largely manual.
Ian Albert says of his process for extracting \emph{Super Mario Bros.} maps:

\begin{quotation}
...these maps were created using FCEU, a free Nintendo emulator for Windows. I used the ROM image for the \emph{Mario/Duck Hunt} version of the game. Screen captures were tediously pieced together in Photoshop. Some text was reproduced using the \emph{Press Start} font by codeman38, which emulates the same font used in \emph{SMB}. Some Game Genie codes were used to make mapping easier... the maps should not be any different due to the use of these codes.~\cite{ian_albert_smb}
\end{quotation}

This workflow requires that a human play through the game, fully test the level (discovering any hidden objects), and periodically take screenshots.
This is only the beginning of the tedious effort involved: at this point, the screenshots must be checked for accuracy (if the level can be changed by player activity they might fall out of sync with each other), and failures here could force the mapper to re-do sections of play.
Next, the screenshots must be stitched together in an external piece of software and any dynamic objects like game characters must be removed.
Finally, all of this must be annotated with semantic information including how the rooms are connected via discrete links!

Every step of this process is time-consuming and error-prone.
Furthermore, due to the ``cheats'' being used, there is no guarantee that the map is truly faithful to the original.
They seem safe for this example, but that does not mean that a different game with different (but similar) cheats might not subtly alter the map (leading to an inaccurate map).
While the map produced by this work supports human interpretation, it requires further human intervention before a machine can effectively process it.
This is due to the image-based format, which requires manual annotation of mechanical properties using an \emph{ad hoc} visual language requiring an understanding of game-specific rules.

For example, Albert uses a visual shorthand of putting a mushroom icon above a block tile to denote that the block in question contains a mushroom.
The actual visual depiction is something that would never be seen in the game (if the mushroom had appeared, the block would not still be a question mark or brick tile), but is read easily by humans.
Similarly, links between maps (between, say, Level 1-1 and the hidden room found by entering the fourth pipe) are depicted with writing on the image. 
In both cases, these annotations would have to be written in a machine-readable form to be processed by a computer, or the computer would have to be ``taught'' how to read the image so as to disambiguate mechanical properties from the spatial map. 

Although many games have been mapped, quality and standards vary significantly from mapper to mapper.
Some mappers produce just the tile backgrounds, while others include the position of all game characters, while others still show hidden mechanical properties.
For some games, there are multiple maps where each describes one such component, but no single map holds all of the relevant information (e.g.\ the \emph{Link's Awakening} maps available at the website Zelda Elements~\cite{linksAwakeningZeldaElements} do not show the contents of treasure chests but do show all characters, while those at VGAtlas~\cite{linksAwakeningVGAtlas} show the full treasure chests but no characters).  

\subsection{Map Extraction}

The shortcomings of manually-produced maps make them unsuitable for some applications---for example, unpopular or hard-to-find games are not likely to have high-quality manually-created maps, and these maps' purely visual representation can make them difficult for automated systems to process.
Extracting maps automatically could circumvent such issues; this is an area of interest for game fan communities as well.

Enthusiasts who want to \emph{modify} a game's maps and other internal data must first be able to identify and extract the game's built-in maps.
This requires detailed, game-specific information about how the map data are stored and encoded.
For games that are routinely modified in such a way, the map formats become a kind of common knowledge; these communities even produce polished software tools to automate the process of viewing and editing maps.
The main games addressed in the present work (\emph{Super Mario Bros.}, \emph{The Legend of Zelda}, and \emph{Metroid}) all have well-understood map formats reverse-engineered from examining source code and memory locations at runtime.

Some games---for example, \emph{Doom} and its successors---define their levels in standalone data files (sometimes wrapped up in larger archives, as in \emph{Doom's} WAD format).
\emph{Doom's} active fan community developed tools to extract these maps and, later, re-pack them to replace the original game's levels.
On the other hand, \emph{Metroid} and \emph{Super Metroid's} levels are only partially defined by bytes of data; the remainder are produced at runtime by an algorithm which is somewhere between decompression and procedural content generation~\cite{metroidLevelFormat}, and these levels can only really be seen accurately by dynamic analysis: watching the game being played over time (or, equivalently, simulating its code to produce the output levels).

This obviously complicates the automatic extraction of maps, but it can still be done on a game-by-game basis with extensive effort.
Static analysis can produce superior maps for those sets of map features where the program's use of the data is well-understood, and once it works for a given game it can work for all games that use the same internal data formats (including modifications of the game).
It can also support games that generate their levels procedurally, since the generation algorithm can be reverse-engineered and fed with different seeds to enumerate possible maps.
Unfortunately, this requires a lot of deep knowledge both of the game's platform and each individual game's machine code.

Some of these limitations can be overcome if the game's runtime memory format is well-understood.
Extant map extraction schemes for \emph{Dwarf Fortress} based on the \emph{dfhack} tool analyze the game's memory structures to pull out complete maps; this hybridizes static and dynamic analysis.  
Conversely, dynamic map analysis can be done just from video as in work by Guzdial~\cite{guzdial2016game}, but it is hard to learn linking structure without control information (see Sec.~\ref{sec:links}).  Guzdial's work associates video frames together into ``chunks'', but it is difficult to know exactly the relationship between these chunks and the levels that they are derived from, as their work finds approximately 47 chunks per level.  Their work also relies upon human annotated spritesheets with game-specific prior knowledge to correctly identify tiles and sprites.

Many approaches to automatic game playing result in the construction of internal maps based on the agent's sensory data.
Rog-o-matic~\cite{rogomatic} constructed three separate maps (an object map, a terrain map, and a room-cycle loop map) which it reasoned over while playing \emph{Rogue}.
Golovin~\cite{golovin} is an interactive fiction playing agent that constructs a map of the world as it travels.
While their game world is depicted as text instead of a graphical representation, it shares some of the same challenges we encounter---namely that a location's depiction might change over time and that multiple locations might share the same depiction.
We note that for these and other game playing agents, map-building is merely an intermediate by-product and not the system's intended output.
Furthermore, these (often special-purpose) approaches typically only require a map to be good enough to guide play, not to be a definitive artifact usable for other purposes.

To summarize the above concerns: automatic map extraction from game data files requires laborious case-by-case reverse engineering.
Doing the same for games that use procedural content generation additionally requires understanding either or both of the game's code and its runtime memory storage formats.
\emph{Mappy} does not require any of this reverse engineering effort, at the same time avoiding the problems of purely video-based techniques by having some knowledge about game \emph{platforms} (in this case the NES), as opposed to specific games.

\subsection{VGLC}

Because gathering game level data has so many complexities, Summerville \emph{et al.}\ assembled the \emph{Video Game Level Corpus} (VGLC)~\cite{summerville2016vglc}.
The VGLC, as of publication time, archives and adapts levels from 12 games into three different map formats.

Some of the highest-quality maps in the corpus were assembled from static analysis (the WAD files for \emph{Doom} and \emph{Doom 2}).
Unfortunately, this approach cannot extend the VGLC very quickly because static level extraction tools are game-specific and difficult to produce.

Half of the games in the overall corpus were added completely by hand-annotation.
The remaining four games' maps were obtained by a mixture of human and computer annotation.
Specifically, Summerville \emph{et al.}\ used template matching to combine a picture of a game map (assembled manually as above) with a human-annotated set of tile types to derive a complete semantic tilemap.

The VGLC proposed three file formats for standardization, of which \emph{Mappy} could be used to generatexs the tile- and graph-based formats.
The tile-based format represents levels as a $W \times H$ matrix, where $W$ is the width and $H$ is the height of the level.
Each element of the matrix is represented as an UTF-8 character.
Along with each level file, there is a \texttt{legend} JSON file that denotes the semantic meaning of each character (e.g.\ \texttt{-} is empty and \texttt{X} is solid).
The graph based format adapts the DOT graph description language to represent rooms (nodes) and doors (edges) between them.
\emph{Mappy} targets the tile-based format for individual rooms and tracks the linkages between rooms using the graph-based format.

\section{Mappy}

\emph{Mappy} is designed to work on games where an \emph{avatar} moves around a large \emph{world} broken up into smaller \emph{rooms}.
This covers significant aspects of a broad class of games including platformers, action-adventure games, and role-playing games.
We based this view of the world on these games' usual composition of four \emph{operational logics}~\cite{wardrip-fruin2005playable,osborn2017refining}: \emph{collision logics}, which describe spaces made up of distinct objects which can touch each other and possibly block each other's movement; \emph{linking logics}, which define larger conceptual spaces including connected rooms and the transit between them; \emph{camera logics}, which account for the fact that the visible part of the world is a window onto a larger contiguous world; and \emph{control logics}, which map e.g.\ button inputs to in-game actions.

Operational logics combine abstract processes (collision detection and restitution, the movement of the player between discrete spaces, the selective drawing of a sub-region of the whole level, or conditional control of the player character) with strategies for communicating these processes to players (tiles and sprites, scrolling or screen-fading to change rooms, continuous smooth scrolling, and ignoring input during cutscene-like segments such as switching rooms).
We find that operational logics provide useful inspiration for knowledge representation and inductive bias; they help structure intuitive observations about how games function in a way that is amenable to automation.
The following sections expand on the leverage we get from operational logics as a knowledge representation.

In its current form, \emph{Mappy} takes as input a playthrough of a game and the game program, then runs an NES emulator on that program and observes the system's state over time.
\emph{Mappy} watches a portion of the screen for changes; this screen rectangle is currently given in advance, but it could be determined automatically in the future.
At each timestep, \emph{Mappy} determines what tiles are visible on the screen, whether the screen is scrolling and if so by how much, and whether the player currently has control over the game (through speculative execution of inputs).
\emph{Mappy} accumulates a map of the current room as the game is played: when \emph{Mappy} sees a new part of a room, it adds those tiles to that room's map.
If a tile in a room changes, \emph{Mappy} also notes that the tile has changed, storing a history of each coordinate's contents over time.
This is important for capturing e.g.\ breakable blocks in \emph{Super Mario Bros.}\ or collapsing bridges.
\emph{Mappy} also watches for cases when the player might be moving between rooms and starts on a new map when the move is complete.
Finally, \emph{Mappy} analyzes the rooms it has seen and suggests cases where two witnessed rooms might actually be the same room so that a human may choose whether to merge them together.

\subsection{NES Pragmatics}

\emph{Mappy} works on NES games because that platform's hardware explicitly defines and supports the rendering of grid-aligned tiled maps (drawn at an offset by hardware scrolling features) and pixel-positioned sprites.
The NES implements this with a separate graphics processor (the \emph{Picture Processing Unit} or PPU) that has its own dedicated memory defining tilemaps, sprite positions (and other data), color palettes, and the \(8 \times 8\) patterns which are eventually rasterized on-screen.
During emulation, \emph{Mappy} can directly read the PPU memory to access all these different types of data; we briefly describe the technical details below (referring the interested reader to~\cite{nesdev}).

Although the PPU has the memory space to track 64 hardware sprites at once, there are two important limitations that games had to contend with: first, each sprite is \(8 \times 8\) pixels whereas game objects are often larger; and second, the PPU cannot draw more than eight hardware sprites on the same \emph{scanline} (screen Y position).
This means that sprites are generally used only on objects that \emph{must} be positioned at arbitrary locations on the screen.

Static geometry, including background and foreground tiles, are not built of sprites but are instead defined in the \emph{nametables}, four rectangular \(32 \times 30\) grids of tile indices; these four nametables are themselves conceptually laid out in a square.
Since the PPU only has enough RAM for two nametables, individual games define ways to mirror the two real nametables onto the four virtual nametables (some even provide extra RAM to populate all four nametables with distinct  tiles).
On each frame, one nametable is selected as a reference point; when a tile to be drawn is outside of this nametable (due to scrolling) the addressing wraps around to the appropriate adjacent nametable.
Note that many game levels are much wider than 64 tiles---the game map as a whole never exists in its player-visible form in memory, but is decompressed on the fly and loaded in slices into the off-screen parts of the nametables as the player moves around the stage.

\emph{Mappy} remembers all the tiles that are drawn on the visible part of the screen, filling out a larger map with the observed tiles and updating that map as the tiles change.
A \emph{Mappy} map at this stage is a dictionary whose keys are a tuple of spatial coordinates (with the origin initially placed at the top-left of the first screen of the level) and the time points at which those coordinates were observed, and whose values are \emph{tile keys}.
A tile key combines the internal index used by the game to reference the tile with the specific palette and other data necessary to render it properly (from the attribute table and other regions of NES memory).
After \emph{Mappy} has determined that the player has left the room (see Sec.~\ref{sec:links}), the map is offset so that the top-left corner of its bounding rectangle is the origin and all coordinates within the map are positive; this is rasterized and output as an image.
We thereby construct the level as it is seen from the perspective of (tile-based) collision logics: the (mostly) static geometry and its (semantically significant) visual appearance over time.

We learn the full history of every tile, rather than committing to its initial or final appearance, for four main reasons.
First, during scrolling, stale tiles are regularly replaced with fresh ones, and in some games this can even happen at the edges of the screen causing visible glitching.
Second, we often fade into or out of rooms (or perform some other animation), and just taking the first- or last- seen tile could lead to unusable maps.
Third, many tiles animate during play (for example, ocean background tiles or glittering treasures).
Finally, the player can interact with many tiles: switches can be flipped, blocks can be broken, walls can be bombed, and so on.
So we must store all the versions of a tile to admit applications like learning tile animations or interactions.
For rasterization and visualization, we generally pick the tile's appearance $25\%$ of the way into its observed lifespan, but this is an arbitrary choice and the generated maps are mainly for human viewing.
A more principled choice might be to take the most common form the tile took during its lifespan.

While in general the nametables are used for terrain and the hardware sprites are used for game characters, there are some exceptions.
Large enemies that do not animate much are often built from background tiles (as in some \emph{Mega Man} bosses and \emph{Dragon Quest} enemies); moving platforms act as terrain but generally must be implemented as sprites.
Objects like movable blocks in \emph{Zelda} or breakable bricks in \emph{Super Mario Bros.}\ are tiles most of the time, but temporarily turn into sprites when interacted with so that they can animate smoothly off of the tile grid.
\emph{Mappy} does not account for these special cases yet.

Because some important level objects are sprites and not tiles, we also hope to learn the initial placements of dynamic game objects in the larger map.
\emph{Mappy} identifies abstract game objects by observing hardware sprites over time using the sprite tracker described in~\cite{summerville2017mechanics}.
This system uses information-theoretic measures to merge adjacent hardware sprites into larger game objects and maintains object identity across time using maximum-weight matchings of bipartite graphs (object identity and positioning in 2D space are natural conclusions to draw from collision logics).
For \emph{Mappy}, we take the first-witnessed position of each object, register those coordinates relative to level scrolling (explained in the next section), and render its constituent sprites into our level maps to capture, for example, that a mushroom pops out of a particular question-mark block.

\subsection{Scrolling}

Although the PPU features hardware scrolling, and (some) of this information persists in the PPU's hardware registers, capturing screen scrolling information is surprisingly subtle.
Games can alter the hardware scrolling registers essentially at any time during rendering, to achieve for example split screens or static menus over scrolling levels (the NES does not support layered rendering, unlike the Super NES).
\emph{Super Mario Bros.}\ and its sequels draw the top part of the screen containing status and score information without scrolling, and then turn scrolling on for subsequent scanlines.
\emph{Super Mario Bros.\ 3} puts status information on the \emph{bottom} of the screen as well, so only a small window of the larger screen scrolls.
These are concrete examples of camera logics, where a portion of the screen is dedicated to a viewport backed by the illusion of a moving camera.
As mentioned above, we register the visible part of the level in a larger tilemap, under the assumption that a rectangular viewport will view a rectangular region of a potentially larger space.

\begin{figure}
  \includegraphics[width=0.22\textwidth]{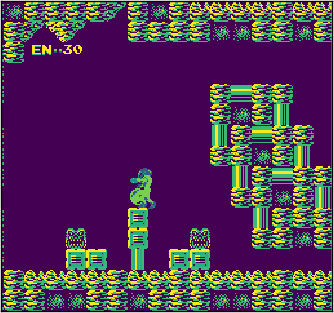}
  \includegraphics[width=0.22\textwidth]{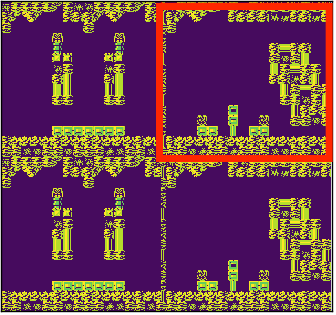}
  \caption{Visible screen registered with PPU nametables.  Note vertical mirroring and horizontal wrapping.} \label{fig:ntas}
\end{figure}

We could have obtained pixel-precise scrolling information by instrumenting the emulator to trace when hardware scrolling state changes, but we wanted to see how far we could get without such interventions to remain as general as possible.
We deploy two techniques, each with their own strengths and weaknesses: a perceptual algorithm based on registering each frame's visual output with the previous frame's and a hybrid approach which registers only the current frame's visual output (converted to grayscale) with the PPU's four nametables to determine which rectangular sub-region of the larger tilemap is being shown (see Fig.~\ref{fig:ntas}).
The former technique can break down with animated backgrounds (for example, waterfalls), while the latter will fail if the perceived scrolling is done mainly by sprites rather than background tiles, as in certain boss fights in \emph{Mega Man 2}---this would also be an issue if we tracked hardware scrolling with the instrumentation described above.
In either case, once \emph{Mappy} has precise scrolling information it can convert coordinate spaces from the subset of tiles drawn on the screen into the frame of reference of the larger map it is assembling.

\subsection{Linked Rooms} \label{sec:links}

\begin{figure}
\centering
\includegraphics[width=0.48\textwidth]{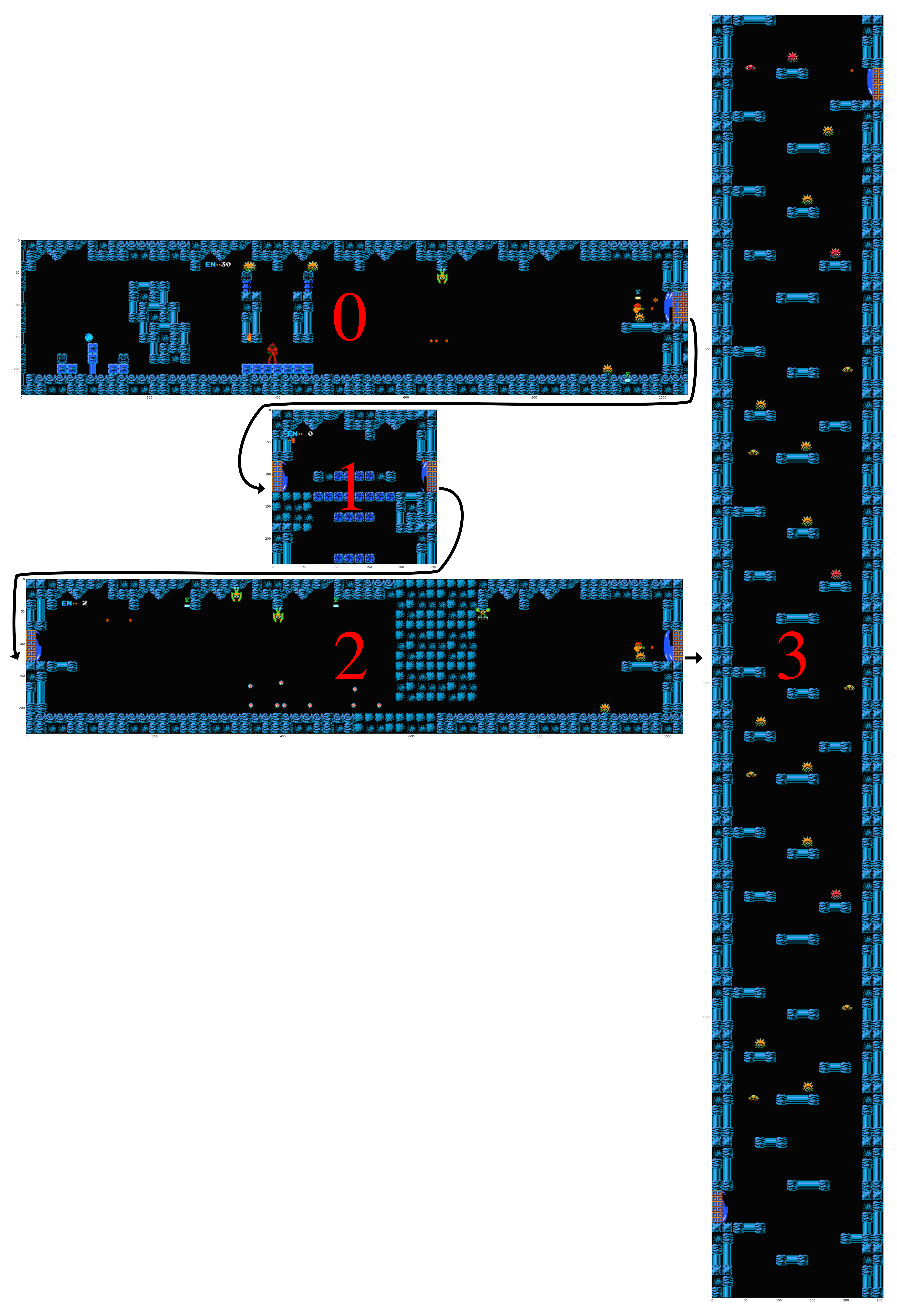}

\caption{The first four rooms from \emph{Metroid}.  Note that we only observed a small portion of room $1$, which is actually another tall vertical corridor.}
\label{fig:metroid}
\end{figure}

In this work we want to learn not only one large tilemap, but the graph structure by which smaller rooms are linked together (game worlds are not in general planar or even Euclidean).
To do this, we need to determine when the player leaves one contiguous space for another.
We consider two main ways in which linking logics communicate room changes to players:
\begin{itemize}
\item Smoothly scrolling between connected rooms
\item Teleporting between rooms
\end{itemize}
The first type of transition is the most common type in \emph{The Legend of Zelda} and \emph{Metroid} (see Fig.~\ref{fig:metroid}).
In these games, when traversing between most rooms the player loses control for a period of time while the screen rapidly but smoothly scrolls completely into the new room.
After the player regains control, they are in a new room.
To test for this type of transition we must know for each frame whether the screen is scrolling and whether the player has control; we already know about scrolling, so we use the savestate features of the emulator to determine whether the player has control.

The central question of player control is: ``Would the world have been different if the player had done something else?''
Because we know the full input sequence we can look a few moves ahead to see how the world will evolve according to the playthrough; we automatically take a screenshot of that state for reference.
Next, we simulate seven possible futures (one for each button besides ``start'') three frames ahead and compare a screenshot taken in each of those eventualities against the reference state.
If these actions produce different outcomes than the reference, then the player must have control at the initial frame.

In many games, some animations enacted by the player implicitly remove player control for some period of time (e.g.\ the fixed length jumps in \emph{Castlevania}), so we have a configurable parameter for how long control must be taken away before counting as a complete loss of control.
Since most room transitions take at least one or two seconds, and most in-game animations remove control for less time than that, this allows for a clean separation of the two causes for losing control.
Of course, it is conceivable that the player does not have control but is not entering a new room, so we stipulate that the screen must also be scrolling while control is lost (and, indeed, that it must have scrolled by at least half the scroll window width/height).
This accounts for freeze frame animations such as when Mario acquires a mushroom and grows or the fanfare that plays when Samus acquires a new item, which show a loss of control but the screen stays stationary.

The second category of spatial transition above places the player in a new room that has little or no visual relation to the previous room, perhaps from descending a staircase or going down a pipe.
We treat these by looking at the overall appearance of the game screen, and if it changes too drastically within a short timeframe we assume that the player has probably teleported to a new room.
This is complicated by game levels that incorporate drastic sudden changes to the visible portion of the tilemap (such as the ``dark storm'' level in \emph{Ninja Gaiden} or Bright Man's stage in \emph{MegaMan 4}), which yield false positives where \emph{Mappy} thinks that it has gone to a different room.
Given the optional room merging discussed below (and the possibility of stronger heuristics which we leave for future work), we do not believe that this is a fatal flaw.

\section{Cleaning Up}
\begin{figure}
\centering
\includegraphics[width=0.48\textwidth]{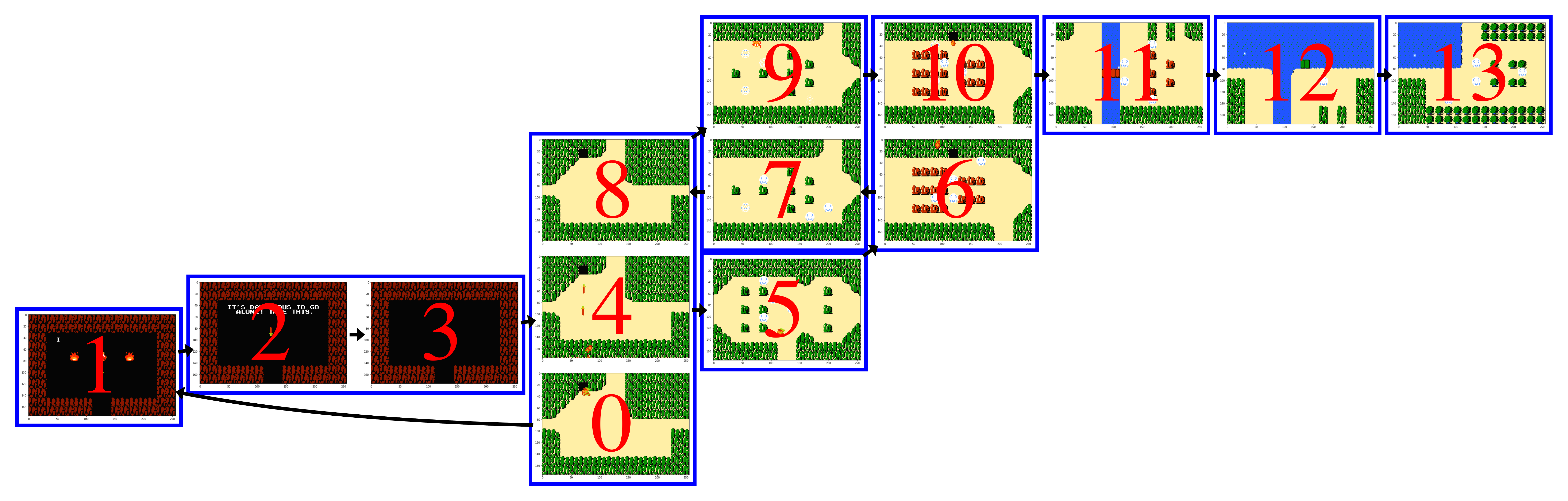}
\caption{Example room linkage detection and room merging step.  Red numbers represent the order of traversal.  Rooms that are believed to be identical are grouped together in blue boxes.  It is up to the user to choose which rooms should or should not actually be merged.}
\label{fig:room_merge}
\end{figure}

At this point, \emph{Mappy} has detected individual rooms and linkages between them, but it assumes that every link leads to a brand new room. 
In most games, at least some links are two-way or converge on the same destinations---most game worlds form a graph and not just a tree.
We could simply merge rooms that look identical to each other, but there are numerous cases where this might fail.
For instance, there are rooms in \emph{The Legend of Zelda} that have identical tilemaps but are in fact different rooms.
There are also instances with more complex mechanics at play: in \emph{Zelda's} ``Lost Woods'', the player moves through a sequence of identical-looking rooms and must use the correct door in each of those rooms or return to the first room in the sequence.
We do not expect to be able to automatically cover all such cases since in the end room connections are defined in opaque game programs and we cannot hope to address every possibility.
We therefore leave it up to a human analyst to select which rooms should or should not be merged.
\emph{Mappy} provides suggestions based on overall similarity; in Fig.~\ref{fig:room_merge}, \emph{Mappy} is largely correct (though it misses the fact that $1$, $2$, and $3$ are the same room) and the final map should consist of the merged rooms $(0, 4, 8), (1, 2, 3), (5), (7, 9), (6, 10), (11), (12), \text{ and } (13)$.

\begin{figure}
\centering
\includegraphics[width=0.48\textwidth]{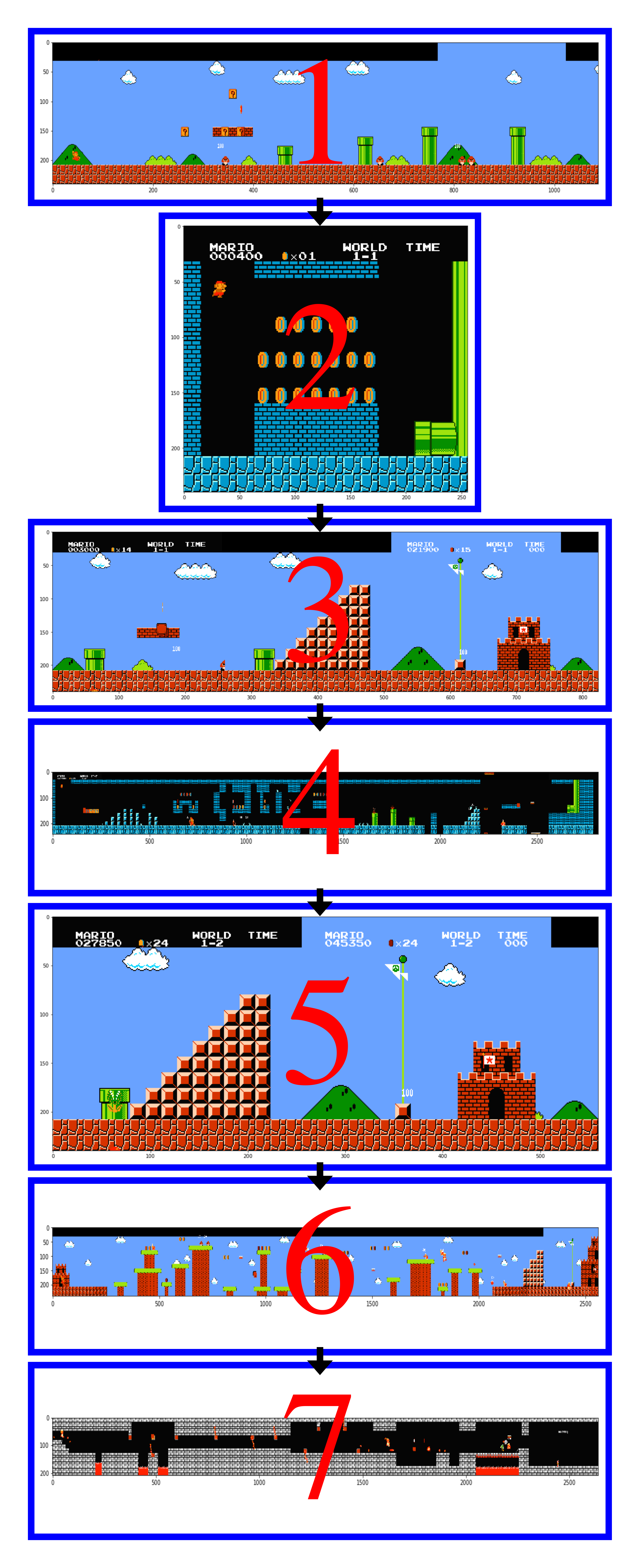}
\caption{The first 4 levels extracted from \emph{Super Mario Bros.}; Level 1-1 is comprised of rooms 1, 2, and 3.}
\label{fig:mario}
\end{figure}

Note that there are important candidate merges \emph{Mappy} does not detect.
For instance, we currently have no method to detect returning to a different part of the same room.
Fig.~\ref{fig:mario} shows an example where the player takes a warp pipe (from room $1$) to a bonus room (room $2$) and then emerges later in the level (room $3$).
The correct map would show that the first room and the third are actually two parts of one larger room; but even a human player must explore multiple paths through the level to determine this.
In the future we intend to use computer vision techniques (e.g.~\cite{mann1993compositing,ward2006hiding}) to merge the results of multiple play traces, so as to be able to fully map segments of a game that are mutually exclusive or are not likely to both appear in the same traversal.

\begin{figure*}
\centering

\includegraphics[width=0.96\textwidth]{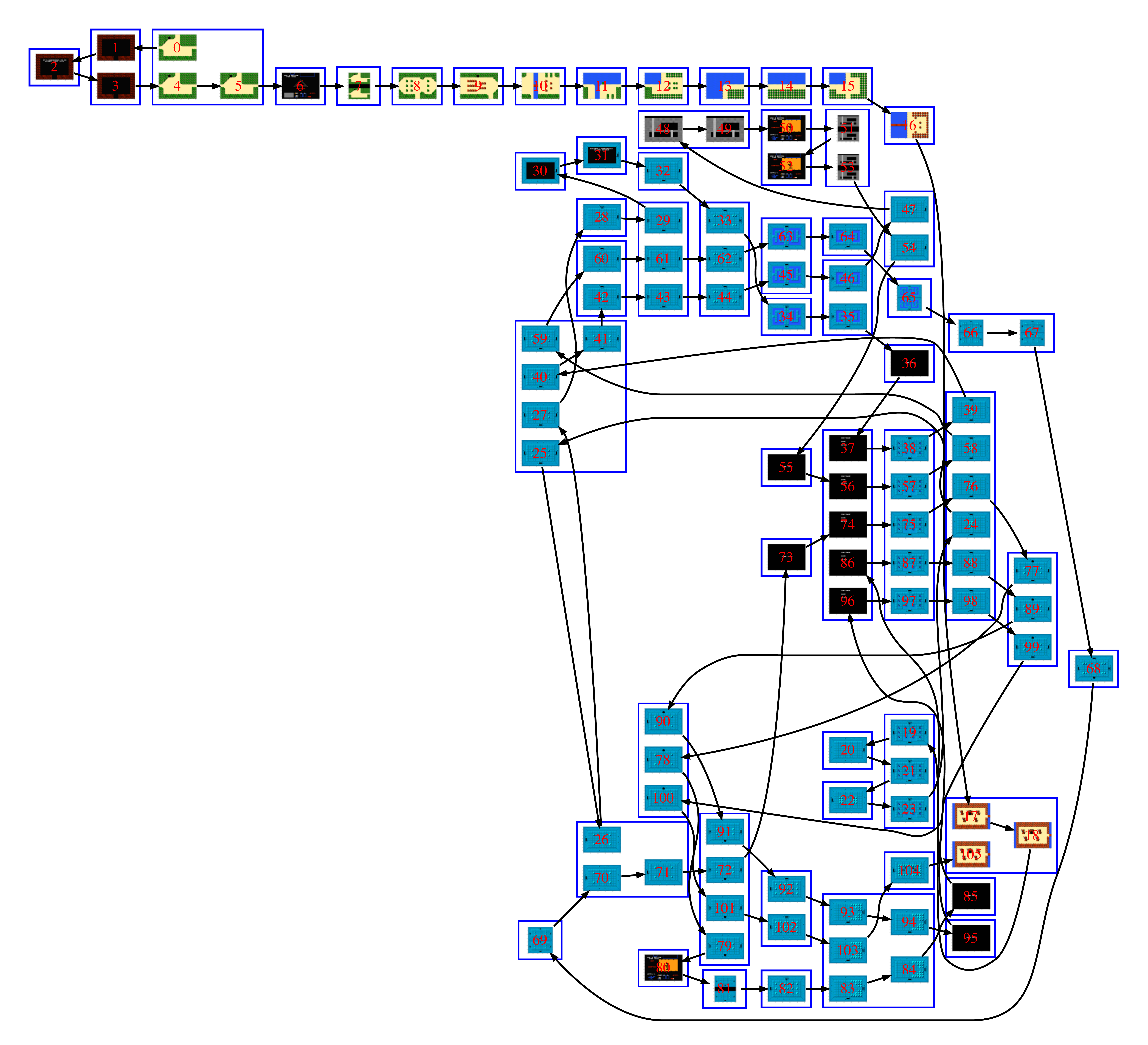}

\caption{\emph{Zelda} through the completion of Dungeon 1.  The player (one of the authors) made numerous mistakes resulting in deaths (the cluster of black screens in the middle) which teleport the player to the beginning of the dungeon.}
\label{fig:zelda_d1}
\end{figure*}

\begin{figure*}
\centering

\includegraphics[width=0.96\textwidth]{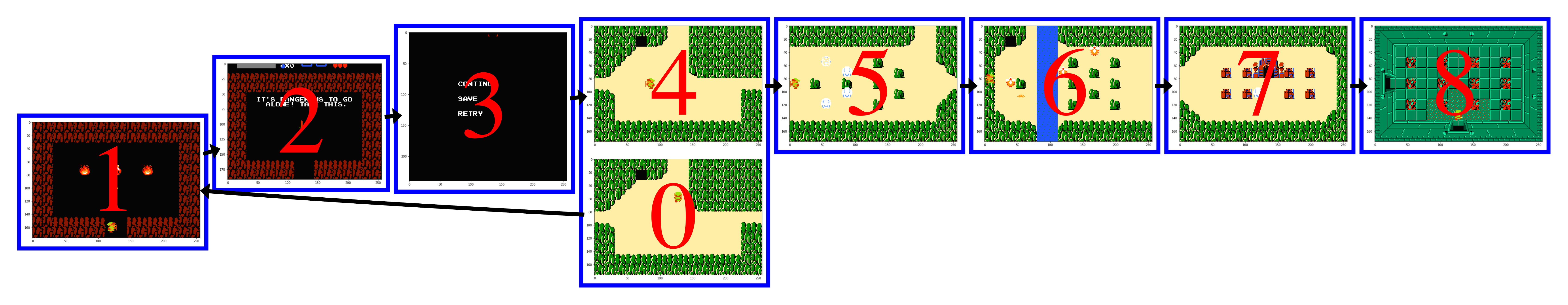}

\caption{\emph{Zelda} up to Dungeon 3, showing a map which is \emph{true} but not \emph{reasonable}.}
\label{fig:zelda_tas}
\end{figure*}

We do not yet have a pleasant user interface for this manual merging process, but we imagine a possible UI closely tied to the visualizations provided in this paper.
A human analyst confronted with a map like Fig.~\ref{fig:zelda_d1} might begin by merging obviously identical rooms according to \emph{Mappy's} suggestions.
Selecting a cluster (e.g.\ $(0,4,5)$) by clicking on it and then hitting return (or double-clicking on the border of the cluster) would collapse these rooms into one, merging the links in and out at the same time.
Next, the two clusters in the top-left corner are actually the same room---but the use of background tiles to render text (combined, perhaps, with loss of control when obtaining the sword) has confused \emph{Mappy}.
Our hypothetical analyst could shift-click to select both those clusters and hit return to merge them into one, and then double-click the blue border of that cluster to combine it into a single room.
Finally, the treasure rooms of most dungeons in the \textit{Legend of Zelda} look nearly identical.
A user could split apart a candidate suggestion by shift-selecting individual rooms of one or more clusters and then hitting return to pull them into their own combined clusters (multiple clusters could simultaneously be combined in the same way, or a combination of clusters and rooms).
In this way, individual treasure rooms or other similar-looking rooms could be kept separate.

The images shown in this paper visualize the first appearance of every game object (with the sprite image it appeared as at that time), but we are really only interested in game objects that are arranged as part of the level.
In \emph{Super Mario Bros.}, we do not want the \texttt{100}-point indicators; in \emph{Metroid} we do not want Samus's shots or the powerups dropped by defeated enemies.
Because figuring out which game objects appear because of the evolution of game rules and which sprites appear as part of the level is not a well-defined problem (consider the mushrooms or stars coming out of question-mark blocks), we leave it to human intuition; However, we intend to resolve this in future work, perhaps leveraging techniques for learning game rules~\cite{guzdial2017ijcai} or interactions~\cite{summerville2017what}.
We do not yet provide a graphical user interface for indicating game objects to exclude from the map; currently we remove unwanted game objects by examining their appearance and then eliminating the corresponding tracked game objects from future processing.

\section{Discussion}

\emph{Mappy} makes several structuring assumptions about games and play, and it is informative to explore where and how these break down.
We have already discussed limitations in scrolling detection and room merging, but there is another important assumption which has not been addressed yet:  \emph{Mappy} implicitly assumes it is observing \emph{natural} play where a human explores the game in the way intended by designers and programmers.
Here it is useful to distinguish \emph{true} maps from \emph{reasonable} ones.
We call maps that accurately reflect game code \emph{true}, even if they are inconsistent with players' expectations of the game's design.
A \emph{reasonable} map is one that matches these expectations but might be unfaithful to the source code (e.g.\ the maps from Zelda Elements mentioned above).
Comparing Figs.~\ref{fig:zelda_d1} and~\ref{fig:zelda_tas} showcases this distinction.
The former is a natural play of the game; but in the latter, a tool-assisted speedrunner utilizes multiple glitches to take an optimal (not at all natural) path.
The first is the so-called ``Up+A'' trick, which causes a soft reset of the game when the player enters the eponymous command on the second controller.
This covers the transitions from $2$ (picking up the sword) to $3$ (the soft-reset screen) to $4$ (the player's initial location at game start).
The second trick is ``Screen Scrolling'', which lets the player leave a screen and re-enter that same screen from the other side.
This is how the player warps over the river in $6$ and (due to collision detection failing when inside an object) passes back through the same wall to room $7$.

All this is allowed by the code of the game, and the \emph{true} map we collected captures the full behavior of that code; of course, this would be inappropriate for many of the use-cases we suggested in the introduction.
A human or AI player would probably want a map that characterizes their understanding of the game world.
A user feeding this map to a machine learning algorithm for design knowledge extraction would likewise want a map that conveys the intended (if not actual) progressions in the game.
It is also interesting to consider that an optimal AI will find such ``secret passages'' while an AI that does semantic hierarchical planning (e.g.\ planning sequences of platforms or rooms to traverse) will probably not.
That said, \emph{true} maps can be valuable to a game creator---particularly for highlighting areas where it differs from a \emph{reasonable} map e.g.\ for detecting bugs or sequence breaks.

As for learning map data proper, one important aspect of links which we currently ignore is that links are embedded \emph{in space}.
In other words, the player usually travels between rooms because the character stepped on a staircase or crossed between rooms.
Right now we do not learn the embedding of the network of links into the tilemaps, but this is important future work.
Notably, the same doorway might take the player to multiple different rooms (if, for example, certain game progress flags have been set) or the same room might be entered on the same side from multiple doorways (as in the Lost Woods).

We see natural future work in extending the set of games which \emph{Mappy} can address both on the NES and on other platforms (including black-box games without the hints from dedicated graphics hardware).
Many of our techniques will transfer readily, but some of the low-level feature extraction must be adapted to work with additional context or on different hardware, perhaps incorporating more techniques from computer vision.
The NES has been productive for our uses, but we do occasionally run into quirks of the hardware that would be avoided with pure computer vision approaches---for example, different games can include custom wiring or even additional memory that our internal tile renderer must handle properly.

We also want to extend \emph{Mappy} to find the scrolling sub-region of the screen automatically.
This might be done by observing which portion of the screen seems to move around smoothly with respect to the whole viewport as the playthrough goes on; at any rate it is extremely important for games like \emph{The Legend of Zelda 2: The Adventures of Link}, where the top-down overworld screen has no status bar while the side-scrolling screens do.
It is especially important to handle game \emph{modes}, including game-over and stage-start screens, battle versus field modes versus menus in role-playing games, and so on.
This is complicated even in \emph{The Legend of Zelda} where the menu activates by smoothly scrolling down and effectively pauses the action on the part of the screen \emph{Mappy} should pay attention to.

As mentioned earlier, \emph{Mappy} ought to analyze several play-throughs of a game to get more complete maps.
We could even borrow techniques from undirected or curiosity-driven search~\cite{stanton2016curiosity,pathak2017curiosity} to reduce the need for human-provided play traces; this could take the form of automatic exploration off of the main branches given by provided traces or even fully automatic search.

As an un-optimized prototype, \emph{Mappy's} runtime performance is acceptable but not great.
Mapping out a minute of play (3600 frames) takes between five and six minutes, mainly due to the expensive scrolling detection and game object tracking.
Obviously this is an area for improvement and we are actively exploring ways both to parallelize \emph{Mappy} and to bring down its constant factors.
One easy way to increase the mapped frames per second (at the cost of missing short-lived tile changes) would be to only make map-related observations every few frames.
Initial experiments here suggest that looking at every second frame is a good compromise that almost doubles the exploration speed without sacrificing too much accuracy; looking at every fifth frame roughly doubles the speed again but the results require postprocessing and cleanup to be made usable.
This time skipping could also be made adaptive, taking longer steps when the visual appearance is not seen to change rapidly.

Finally, we hope to track the \emph{provenance} of \emph{Mappy's} conclusions about maps.
In other words, we would like to identify for a given link, map, game object position, or other observation what game state (or sequence of game states) witnessed that fact.
This could help improve the quality of our conclusions---in some cases we may want to interpret a screen transition as indicating either a change in room \emph{or} merely a menu popping up, and tracking why we believe one or the other conclusion seems useful for optimally resolving the ambiguity.
Provenance also could improve the experience of merging rooms: being able to click and load up a pair of similar rooms in an emulator could help an analyst decide if they are indeed the same room.
Moreover, this magnifies the utility of search and retrieval over concepts like level fragments, linking structures, or which sprites appear in which rooms.
We believe a database that admits querying for e.g.\ infinitely-looping hallways or Lost Woods-style trick dungeons could be useful for scholars of digital games as well as for data-driven PCG, and \emph{Mappy} points the way to building tools like that at the same time it helps improve the coverage of the VGLC.

\bibliographystyle{ACM-Reference-Format}
\bibliography{mappy} 

\end{document}